# The Three Ensemble Clustering (3EC) Algorithm for Pattern Discovery in Unsupervised Learning

Debasish Kundu
Founder TUNL (an AI startup)

debkundu@tunl.tech
debkundu@gmail.com

**ABSTRACT**

This paper presents a multiple learner 'Three Ensemble Clustering (3EC)' algorithm that classifies unlabeled data into quality clusters as a part of unsupervised learning. It offers the flexibility to explore the context of new clusters formed by an ensemble of algorithms based on internal validation indices.

It is worth mentioning that the input data set is considered to be a cluster of clusters. An anomaly can possibly manifest as a cluster as well. Each partitioned cluster is considered to be a new data set and is a candidate to explore the most optimal algorithm and its number of partition splits until a predefined stopping criteria is met. The algorithms independently partition the data set into clusters and the quality of the partitioning is assessed by an ensemble of internal cluster validation indices. The 3EC algorithm presents the validation index scores from a choice of algorithms and its configuration of partitions and it is called the Tau(Γ) Grid. 3EC chooses the most optimal score. The 3EC algorithm owes its name to the two input ensembles of algorithms and internal validation indices and an output ensemble of final clusters.

Quality plays an important role in this clustering approach and it also acts as a stopping criteria from further partitioning. Quality is determined based on the quality of the clusters provided by an algorithm and it's optimal number of splits. The 3EC algorithm determines this from the score of the ensemble of validation indices. The user can configure the stopping criteria by providing quality thresholds for the score range of each of the validation indices and the optimal size of the output cluster. The users can experiment with different sets of stopping criteria and choose the most "sensible" group of quality clusters.

.

# 1. The Renewed Imperative for Unsupervised Learning

Some changes caused by disruptions are easily manifested and easy to track while there are others which do not manifest easily and lie undetected as insignificant anomalies. Machine Learning Ops (Operations) mandates EDA (exploratory data analysis) but it is important to review new data from an unlabeled, unsupervised perspective because new anomalies may be the new normal. Examples of some of these disruptions that are impacting us daily are the human behavioral changes because of the Covid-19 pandemic, the cybersecurity space hackers who are continually striving to invent new techniques, new variants of viruses that frequently emerge, impact of climate change and many more. All these changes require pattern and anomaly detection for data from all relevant channels. Thus, the main concern in the clustering process of unsupervised learning is to reveal the organization of patterns into "sensible" groups, which allow us to discover similarities and differences, as well as to derive useful conclusions about them.

Clustering is a knowledge discovery process that is one of the most important methods of unsupervised learning. The clustering problem is about partitioning a given data set into groups (clusters) such that the data points in a cluster are more similar to each other than the points in different clusters (Guha et al., 1998). Clustering may be found under different names in different contexts (Zimek, A., Vreeken, J), such as unsupervised learning (in pattern recognition), numerical taxonomy (in biology, ecology), typology (in social sciences) and partition (in graph theory) (Theodoridis and Koutroubas, 1999).

The increase in unknown patterns and partitions evident in data from existing and new data sources have led to the invention of new clustering algorithms or variations of existing ones.

Clustering has the following problems in unsupervised learning:

(1) **The 'right' number of clusters is not known:** It is very difficult to know the 'right' number of clusters in unsupervised learning where no observation comes with any label. In fact, the `right' number of clusters in a data-set often depends on the scale at which the data is inspected, and sometimes equally valid (but substantially different) answers can be obtained for the same data (Chakravarty and Ghosh, 1996)[4].

A very interesting example of this observation is in geophysical data integration (Kenneth L. Kvamme, et al, 2019)[5]. A k-means cluster analysis of six geophysical dimensions at Army City yields a number of insights. At k = 2, the class solution divides the region into classes representing "archaeological" anomalies versus background. We get different results of k at values of 3 and 4. Finally, k = 6 best represents important anomaly classes ranging from brick and concrete floors, to walls, burned features, street gutters, and pipelines (Kenneth L. Kvamme, et al, 2019).

(2) **Clustering methods perform differently on the same dataset**: The shape and centroids of clusters formed by different clustering methods are different. As such their performance varies according to different situations. For example, the popular k-means algorithm performs miserably in several situations where the data cannot be accurately characterized by a mixture of k Gaussians with identical covariance matrices (Karypis et al., 1999).

(3) **Quality of the clustering is difficult to determine with unlabeled data:** While there are validity indices that provide the quality of partitioning based on information intrinsic to the data alone, it is not adequate to provide the optimal clusters from a domain perspective.

(4) **Methods will need change with increasing diversity of data:** With increasing veracity, variability and diversity of data, trends and patterns change. So, clustering is dynamic and the same method which helped to successfully transition from an unsupervised learning to a semi-supervised learning may not be adequate in partitioning a new set of data. For example the language of expression changes as we move from one generation to another and so does the engineering of unsupervised learning.

## 2. Motivation for the Ensemble Approach

The notion of integrating multiple data sources and/or learned models is found in several disciplines, for example, the combining of estimators in econometrics, evidence in rule-based systems and multi-sensor data fusion. A simple but effective type of such multi-learner systems are ensembles , wherein each component learner (typically a regressor or classifier) tries to solve the same task (Sharkey 1996; Tumer & Ghosh 1996).
Cluster ensembles provide a tool for consolidation of results from a portfolio of individual clustering results (Strehl and Ghosh, 2002 ).

(1) A bridge between Unsupervised to Semi-supervised and Supervised learning: Classification and Regression problems have several well-defined approaches to implement the ensemble of findings of different algorithms. That is in the space of supervised learning. In the 'Three Cluster Ensemble (3EC)' approach we are exploring clustering of unlabeled data. The cluster results of different algorithms are compared based on the quality of the resulting partitioning as indicated by an ensemble of cluster validity methods

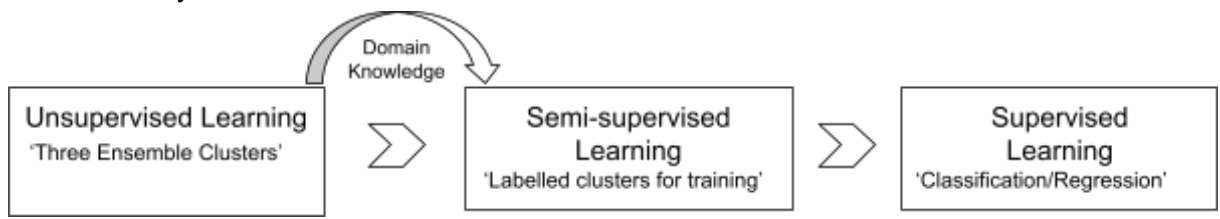

In the case of mental health behavioral modelling the exact same approach has been used (Bhalaji Natarajan et al, 2017).The responses obtained from the target group for the designed questionnaire were first subject to unsupervised learning techniques. The labels obtained as a result of clustering were validated by computing the Mean Opinion Score. These cluster labels were then used to build classifiers to predict the mental health of an individual (Bhalaji Natarajan et al, 2017).

(2) An ensemble of algorithms and validity indices that helps to choose the two important aspects of the clustering. The 'best-fit' algorithm to be used for partitioning of an input data matrix and its configuration of the total number of partitions.

(3) Improves the quality and robustness of results: One can also consider the use of cluster ensembles for the same reasons as classification ensembles, namely to improve the quality and robustness of results. For classification or regression problems, it has been analytically shown that the gains from using ensemble methods involving strong learners are directly related to the amount of diversity among the individual component models (Krogh and Vedelsby, 1995, Tumer and Ghosh, 1999). One desires that each individual model be powerful, but at the same time, these models should have different inductive biases and thus generalize in distinct ways (Dietterich, 2001). So it is not surprising that ensembles are most popular for integrating relatively unstable models such as decision trees and multi-layered perceptrons (Strehl and Ghosh, 2002).

# 3.  The 'Three Ensemble Clustering (3EC)' Algorithm

This paper presents a multiple learner 'Three Ensemble Clustering' algorithm that reveals the organization of patterns inherent in data as quality clusters. Firstly a set of clustering algorithms (being called an ensemble of algorithms) partitions the data into different clusters based on the logic unique to each clustering algorithm. As such an anomaly considered to be insignificant by one algorithm may be presented as a pattern by the other. Secondly, a set of internal validation indices (being called an ensemble of indices) evaluates the clusters and finds the best-fit cluster through a validation function $\Gamma$ (Tau). This ensemble of validation indices is core to the quality aspects of clustering. Thirdly, we get a final ensemble of quality clusters where each cluster may have its origin to a different clustering algorithm and it cannot be further partitioned because it will breach the quality standards.

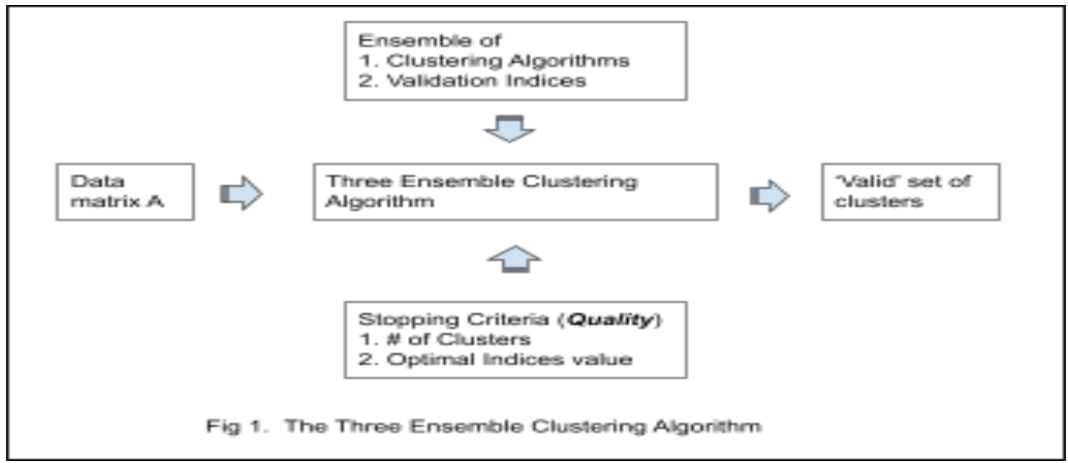

Fig 1. The Three Ensemble Clustering Algorithm

Quality plays an important role in this clustering approach and it also acts as a stopping criteria from further partitioning. Quality is determined based on the optimality of validation index scores. The user provides input quality thresholds for each of the internal validation indices. These thresholds and a user provided optimal length of the cluster form the stopping criteria. The validation function Γ chooses the algorithm and it's configuration of the total number of cluster splits (called Φ-c pair later) that is a best fit for the input data. It presents a Tau-grid which is the score across different Φ-c pairs. The user can experiment with different sets of stopping criteria and choose the best fit 'quality clusters' to explore the domain knowledge and choose the most "sensible" group of 'quality clusters.'

In effect, for any input data the 3EC algorithm will repeat splitting the data into further partitions and these partitions into further partitions until any of the stopping criteria is met. The validation index Γ decides whether it is split by the same partitioning algorithm or a different one. Once any of the stopping criteria is met then the validation function Γ does not split any further. It means that further splitting will lead to suboptimal clusters. The input data matrix to Γ then becomes a part of the final output cluster ensemble.

The algorithm owes its name to an ensemble of algorithms, an ensemble of validation indices that yields an ensemble of 'valid' clusters. Hence 3EC.

# 4) The Ensembles in Three Ensemble Cluster Algorithm

There are two configuration ensembles and an output ensemble of the final cluster in the Three Ensemble Cluster Algorithm.

(1) Ensemble of Algorithms: Different algorithms engage different techniques to clustering based on their approach to the dimensionality and type of data, the separation within the partition and between partitions, the fundamental approach to grouping (like hierarchical versus partition and other types) [ Jain, et al 1999][12].
In our notations, Φ denotes an algorithm which is one of the set of algorithms { Φ$_1$, Φ$_2$, ...Φ$_g$ }. So, Φ$_1$ may be a K-means algorithm, while Φ$_2$ may be a hierarchical clustering algorithm and so on. This ensemble is a user-defined input configuration.

More on clustering techniques in (Jain, Murty, Flynn 1999)[12] and (Halkidi, et al 2001)[13]. There are also autoencoders which also facilitate partition of unsupervised learning.

(2) Ensemble of Validity Indices: The validity indices help in the evaluation of clustering results that best fit the underlying data. In general, clusters should be similar and well separated((Halkidi, et al 2001)[13]. We consider the validity index V which is one of the validity indices {V1,V2,...Vv}. This ensemble is a user-defined input configuration.

More on clustering techniques in (Jain, Murty, Flynn 1999)[12] and (Halkidi, et al 2001)[13].

(3) Ensemble of Clusters: This is the output of the 'Three Ensemble Clustering.' It is a series of final partitions that get collected because no more partitioning is possible. P is the final set of quality partitions $\{P_1, P_2,..,P_l\}$ where $P_x$ is a data matrix x × p and $\sum_{x=1}^{l} P_x = A$ (the input data matrix to the Three Ensemble Clustering(3EC).

# 5. The Stopping Criteria

As mentioned in section 3, the algorithm continues to partition the output cluster until the stopping criteria is met. The validation function Γ enforces the stopping criteria if any of the following conditions are true.

1) μ, the minimum size of a cluster: Often a partitioning algorithm pair Φ-c is forced to make c clusters because it is forced by the configuration parameter of c number of partitions. As such it is likely that some clusters may have only a few objects. That may or may not be correct. The definition of 'few' is also subjective based on the size of the data and the frugal knowledge of the domain at the discovery phase. It is an experimental choice that helps in deliberation of the final ensemble of patterns.

If the Φ-c pair results in a partitioning where the output cluster has less than μ number of objects then that partitioning is discarded. Φ-c is suboptimal.

2) λ,the stopping value of clustering index $V$: The clustering index ensemble {V1, V2, ….Vv} determines the quality of the partitioning based on internal criteria, which is information intrinsic to the data alone. As an example, V1 may be the Davies-Bouldin index while V2 may be the Calinski-Harabasz index and Vv may be a different index. A stopping criterion $\lambda x$ is a value of the index Vx that determines the threshold of optimal quality. If this threshold is breached then the pair Φ-c is considered suboptimal and the partitioning is discarded.

As an example, if V1 is the Calinski-Harabasz index and $\lambda_1$ has a value of 0.3 then a value lower than 0.3 will be considered to be poor partitioning and as such partitioning will be stopped.

# 6. The Validation function 'Tau'

Let the data matrix A = {a1, a2, a3, , an} denote a set of objects. Each row is an observation corresponding to an individual point/object and each column represents a variable observed for all the individuals. There are n observations and p variables. The size of matrix A is n x p.

The 3 Ensemble Cluster (3EC) algorithm partitions an input data matrix A based on a validation function Γ that is both the discriminator and executor for the best-fit quality partitioning. If Γ determines that the input data matrix can be partitioned into quality clusters then it chooses the best-fit algorithm (Φ) and its 'number of clusters' (c) configuration for partitioning. The algorithm and its configuration pair of the number of clusters is called the algorithm and number of cluster pairs, represented as Φ-c. This continues until there is no cluster left that can be further partitioned into quality clusters. These clusters ($P_1$, $P_2$,..,$P_n$) that cannot be further partitioned for breach of quality standards, create the final output cluster ensemble ($P_1$, $P_2$,..,$P_n$) such that Px is a data matrix, $x \times p$ and $\sum_{x=1}^{n} Px = A$.

# 7. Symbols and Notations

| A (Input) | Input data matrix A of n rows and p columns |
|---|---|
| P (Output) | P is the final ensemble of 'quality' partitions ($P_1$, $P_2$,..,$P_n$) where Px is a data matrix $x \times p$ and $\sum_{x=1}^{n} Px = A$. |
| Dm, D | Dm is the most current list of partitions ready to be evaluated for valid partitioning. D is the current partition being evaluated for valid partitioning by the validation function Γ |
| **Ensembles** | |
| Ensemble of Algorithms | Ensemble of Algorithms: Algorithms take different techniques to clustering based on their approach to the dimensionality and type |
| (Input Configuration) | of data, the separation within the partition and between partitions, the fundamental approach to grouping (like hierarchical versus partition and other types) [ Jain, et al 1999].<br>In our notations, $\Phi$ denotes and algorithm which is one of the set of algorithms $\{\Phi_1, \Phi_2, ...\Phi_g\}$. So, $\Phi_1$ may be a K-means algorithm, while $\Phi 2$ may be an hierarchical clustering algorithm and so on. |
| V: Ensemble of Internal Validation Indices - V<br><br>(Input Configuration) | V is a cluster validation index (internal), intrinsic to the information in the data matrix. V is one of $\{V_1, V_2, ....V_v\}$ - an ensemble of validation indices. V1 may be the Davies-Bouldin index while V2 may be the Calinski-Harabasz index and Vv may be a different index. |
| P (Output) | P is the final ensemble of 'quality' partitions $(P_1, P_2,..,P_n)$ where Px is a data matrix x × p and $\sum_{x=1}^{n} Px = A$. |
| **Stopping Criteria** | |
| β | β is a stopping criterion for quality partitioning. It denotes the minimum size of a cluster that represents quality clustering. If a partitioning results in a partition $P_x$ where the size of the partition $P_x$ is less than β then that partitioning is considered invalid. |
| λ | λ is a stopping criterion for quality partitioning. It is one of $\{\lambda_1, \lambda_2, .....\lambda_v\}$ which corresponds respectively to the stopping criterion of validation indices $\{V_1, V_2, ....V_v\}$. A stopping criterion $\lambda_x$ is a value of the index $V_x$ that is determined to be of poor quality and it defines the threshold of optimal quality.<br>As an example, if $V_1$ is the Calinski-Harabasz index and $\lambda_1$ has a value of 0.3 then a value lower than 0.3 will be considered to be poor partitioning and as such partitioning will be stopped. |
| **The Validation Function Γ** | |
| Φ-c pair OR a-c pair (OUTPUT) | An algorithm partitions a data matrix into c partitions based on it's input configuration of the total number of clusters (c). The quality of partitioning depends both on the algorithm and it's configuration of the number of partitions. Partitioning can also depend on other parameters but in this algorithm we have considered partitioning primarily on the input parameter of the total number of clusters. So the algorithm and it's configuration of the number of clusters are inseparable and it is called the algorithm and number of clusters pair. It is represented either as Φ-c or a-c. |
| $\varphi_b$ | Φ is a partitioning algorithm. $\Phi_b$ is the best-fit algorithm from an ensemble of partitioning algorithms $\{\varphi_1, \varphi_2,..\varphi_g\}$. number of |
| | algorithms. $\Phi_b$ is an output of the validation function $\Gamma$. |
| $\mu$ | $\mu$ is the value of c in the best-fit a-c pair which produces the best quality partitioning. |
| $\Gamma$ | The validation function $\Gamma$ evaluates an input data matrix for partitioning within the defined thresholds of the stopping criteria ($\lambda$ and $\beta$) and it outputs the best (a,c) pair - ($\Phi_b$, $\mu$). If the stopping criteria is met then no partitioning happens and $\Gamma$ returns (Null, -1). |
| $\omega$ | $\omega$ is the optimal function for validation indices. Optimality may either mean that a lower score means better quality or a poor quality.<br>For instance in the scikit-learn learning library, the optimality of the Calinski-Harabasz index score is defined as 'the score is higher when clusters are dense and well separated, which relates to a standard concept of a cluster.' Whereas for the Davies-Bouldin Index ' a lower Davies-Bouldin index relates to a model with better separation between the clusters.' |
| Valid Partition | A Valid Partition is a partition which meets the quality thresholds. It splits into partitions $\{S_1, S_2,...S_n\}$ where any partition has a size greater than $\beta$ and the validation indices $\{V_1, V_2, ....V_v\}$ have a score beyond the threshold of $\{\lambda_1, \lambda_2, .....\lambda_v\}$. The significance of being beyond the threshold is that it is lower than $\lambda$ if a particular validation index V indicates better partitioning with lower values of score and higher than $\lambda$ if the vice-versa is true.<br><br>A partition P is considered valid for partitioning if $\Gamma(P)$, which outputs $\{\varphi_b, \mu\}$ has a value of $\mu > 1$. |
| $\Gamma$ | The validation function $\Gamma$ evaluates an input data matrix for partitioning within the defined thresholds of the stopping criteria ($\lambda$ and $\beta$) and it outputs the best of the algorithm and number of cluster pairs ($\Phi_b$, $\mu$). If the stopping criteria is met then no partitioning happens and $\Gamma$ returns (Null, -1). |
| Cmax | A clustering algorithm $\Phi$ is evaluated for its best performance across a range of cluster numbers. Cmax is the pre-defined maximum number of clusters in that range. |
| $\omega$ | $\omega$ is the optimal function for validation indices. Optimality may either mean that a lower score means better quality or a lower score means poor quality.<br>For instance the scikit-learn learning library in the case of Calinski-Harabasz index considers that 'the score is higher when clusters are dense and well separated, which relates to a standard concept of a cluster.' Whereas in the case of the |
| | Davies-Bouldin Index ' a lower Davies-Bouldin index relates to a model with better separation between the clusters.'<br><br>ω has 3 representations:<br>    (1) ω - optimality of a given validation index<br>    (2) $ω_c$ - optimal ζ value on the output of a partitioning algorithm varying the number of clusters from 2 to $C_{max}$<br>    (3) $ω_{ac}$ - optimal ζ value on a set of $ζ_{ac}$ values |
| ζ | ζ is the score for a validation index V. The validation index V gives a score ζ for the partitioning done by an algorithm Φ into c number of clusters, which is called the Φ-c pair. The representations are as follows.<br>    (1) For a given $V_k$, $φ_a$ - c pair, ζ score is represented as $[ζ_k]_{ac}$ score<br>    (2) $[ζ_k]_{ac} = V_k[φ_a(D)]^c$<br>    (3) $[ζ_k]_{ac'}$ (most optimal $[ζ_k]_{ac}$) = $ω[[ζ_k]_a]_2^{C_{max}}$<br>    (4) $[ζ_k]_{a'c'}$ = most optimal $ζ_k$ across all a-c pairs (all algorithms) |

# 8. The Mathematical pseudo-code for the Algorithms

8.1) The Three Ensemble Clustering algorithm is presented as follows.

    STEP 1: Configure the Ensembles and Cmax

    STEP 2: Initialize Cmax and the STOPPING CRITERIA β and (λ1,λ2 ,.....λv ).

    STEP 3: Dm = (A)

    STEP 4: D = the first or next element from Dm

    STEP 5: Delete D from Dm

    STEP 6: {φb, µ} = Γ(D)

    STEP 7: if (µ == -1)
            append D to P ## P is final Cluster Ensemble
      else
            append Dm with partitions from φb [D]µ

STEP 8: if (Dm is NOT Empty) go to STEP 3

STEP 9: return P

## 8.2) The Validation function Γ

for k = 1 to v # total number of indices
    /* find a-c (algo-cluster) pair that gives most optimal score */
    for a = 1 to p # p = total number of algos
        for c = 2 to Cmax # cluster sizes
            $[\zeta_{ka}]^c = V_k[\varphi_a(D)]^c$ # ζ is v × p× Cmax vector
            **STEP 1:** For a given $V_k$ and $\varphi_a$, record $[\zeta_k]_{ac}$ score
            **STEP 2:** Eliminate the ones that breach Quality thresholds
        Get ζ array for for $V_k$ index on $\varphi_a$ varying c to Cmax
    /* out of loop c */
    $[[\zeta_{ka}]^c] = \omega_c [[\zeta_k]_a]_2^{Cmax}$ /* find c that gives most optimal $[\zeta_{ka}]$ */
    **STEP 3:** For a given $V_k$ and $\varphi_a$, find the a-c' pair for the most optimal $V_k$
  /* out of loop a */
  $[\zeta_k]_{a'c'} = \omega_{ac} [\zeta_{kac'}]_1^p$ /* find a-c' that gives most optimal $[\zeta_k]$ */
  **STEP 4:** For a given $V_k$, find the a'-c' pair that gives most optimal $V_k$
    Stopping-criteria check:
      If $[\zeta_k]_{a'c'}$ is less optimal then $\lambda_k$, discard algorithm
      Move to next algorithm
    Find a'-c' pair which gives the most optimal $\zeta_{kac'}$
    For a given $V_k$, find the a'-c' pair that gives most optimal $V_k$
    Append Optimal a-c list
/* out of loop k */
**STEP 5 :** If none of the $[V_k]_1^v$, have any valid a'-c' pair
        φ = NULL,  μ = -1
    ELSE
        φ = most voted algorithm (a'-c' pair)
        μ = corresponding cluster numbers of φ

There may be a few occasions where there is a tie on the most voted algorithm. That solution is beyond the scope of the current paper because that in itself is a treatise that needs a separate paper.

# 9. Here is an example of how the validation function Γ works.

| | Algorithm for Γ | | | | |
|---|---|---|---|---|---|
| Stopping Criteria | Size of cluster (β) | | 10 | 10 | 10 |
| The Γ Fn : a-c' pair in green, breached ones in red | | | | | |
| | | | Algorithms | | |
| | Indices(Vk) | Clusters(C) | $\zeta[A1(D)]^c$ | $\zeta[A2(D)]^c$ | $\zeta[A3(D)]^c$ |
| Step1: For a given Vk and φa, record [ζka]c score | V1 (*higher* score $\zeta_1$ is **Optimal**) (**λ1 = 0.3**) | 2 (C1) 3 (C2) 4 (Cmax) | 0.6 0.4 0.25 (λ1 breach) | **0.7** 0.6 0.5 | 0.6 0.5 0.4 (β breach) |
| Step 2: For a given Vk and φa, eliminate the ones that do not meet the quality thresholds | V2 (*higher* score $\zeta_2$ is **Optimal**) (**λ2 = 275**) | 2 (C1) 3 (C2) 4 (Cmax) | **752** 609 549 | 751.9 608 548 | 636 631 267 (λ2 breach) |
| Step 3: For a given Vk and φa, find the a-c' pair that gives most optimal [ζka]c' score | V3 (*lower* score $\zeta_3$ is **Optimal**) (**λ3 = 0.75**) | 2 (C1) 3 (C2) 4 (Cmax) | **0.58** 0.64 0.65 | 0.59 0.64 0.65 | 0.63 0.72 0.8 (λ3 breach) |
| The Γ Fn : a'-c' pair for every Validation score | | | | | |
| Step 4: For a given Vk, find the a'-c' pair that gives the most optimal Vk ([ζk]a'c') | V1 (higher score is **Optimal**) | 2 (C1) 3 (C2) 4 (Cmax) | | A2-2 | |
| | V2 (higher score is **Optimal**) | 2 (C1) 3 (C2) 4 (Cmax) | A1-2 | | |
| | V3 (lower score is **Optimal**) | 2 (C1) 3 (C2) 4 (Cmax) | A1-2 | | |
| The Output of the Validation Function Γ | | | | | |
| Step 5 (OUTPUT): If there is no optimal a'-c' pair then make φ NULL and μ is -1 ELSE φ = most voted algorithm μ = corresponding cluster numbers of φ | | | | **(A1, 2)** | |

# 10. Case-Study

The validity of a cluster lies in the eye of the business domain expert. The 3EC algorithm has been used in several discovery projects; however, it requires a good domain understanding to interpret the newly found clusters. As such, a simple case study of one of the most used dataset is being presented - the IRIS data set

The IRIS data set is a labelled dataset but we have removed the labels and explored 3EC for partitioning the data into clusters.

## 10.1 Initial Configuration of 3EC:

The initial input configuration is presented below. The explanation of the individual algorithms in the Ensemble of Algorithms and the individual Validation Indices (internal) in the Ensemble of Indices is beyond the scope of this paper.

| Ensembles | |
|---|---|
| A | Iris dataset (without labels) |
| Ensemble of Algorithms | $\Phi_1$ = K-means <br> $\Phi_2$ = Agglomerative Clustering (Agglo) <br> $\Phi_3$ = Spectral Clustering |
| V (Internal Validation Indices) | $V_1$ = Silhouette index (S) <br> $V_2$ = Calinski-Harabasz index (C) <br> $V_3$ = Davies-Bouldin index (D) |
| Cmax | 3. So the range of the clusters are {2,3,4}. |
| Stopping Criteria | |
| β | 40 |
| λ | $\lambda_1$ (< 0.45 ) <br> $\lambda_2$ (< 500 ) <br> $\lambda_3$ (> 0.80 ) |

## 10.2 Execution:

### 10.2.1 Determining the Tau grid at the input cluster A is as follows

| Models/Algos | K-Means | | Agglomerative | | Spectral-NN | |
|---|---|---|---|---|---|---|
| | Silhouette score Index Analysis | | | | | |
| Rank | Clust-Size | Score | Clust-size | Score | Clust-size | Score |
| 1 | 2 | 0.682 | 2 | 0.687 | 2 | 0.687 |
| 2 | 3 | 0.553 | 3 | 0.554 | 3 | 0.554 |
| 3 | 4 | 0.498 | 4 | 0.489 | 4 | 0.494 |
| | Calinski-Harabasz Index Score Analysis | | | | | |
| Rank | Clust-Size | Score | Clust-size | Score | Clust-size | Score |
| 1 | 3 | 561.63 | 3 | 561.63 | 3 | 556.88 |
| 2 | 4 | 530.71 | 4 | 530.77 | 4 | 523.1 |
| 3 | 2 | 513.92 | 2 | 513.92 | 2 | 502.82 |
| | Davies-Bouldin Index Score Analysis | | | | | |
| Rank | Clust-Size | Score | Clust-size | Score | Clust-size | Score |
| 1 | 2 | 0.404 | 2 | 0.383 | 2 | 0.383 |
| 2 | 3 | 0.662 | 3 | 0.656 | 3 | 0.658 |
| 3 | 4 | 0.78 | 4 | 0.795 | 4 | 0.793 |

*Fig. 1: The Tau grid for input cluster = A*

The $\Gamma$ function outputs the $\varphi_b$ - $\mu$ pair as K-means and 2 respectively. There are two valid partitions S1 and S2. The stopping criteria is met for S1 which means it cannot be further split (because further partitioning creates clusters of length less than the stopping criterion β. So, S1 becomes the first final cluster P1 of the ensemble of final clusters. The partitioning is as follows:

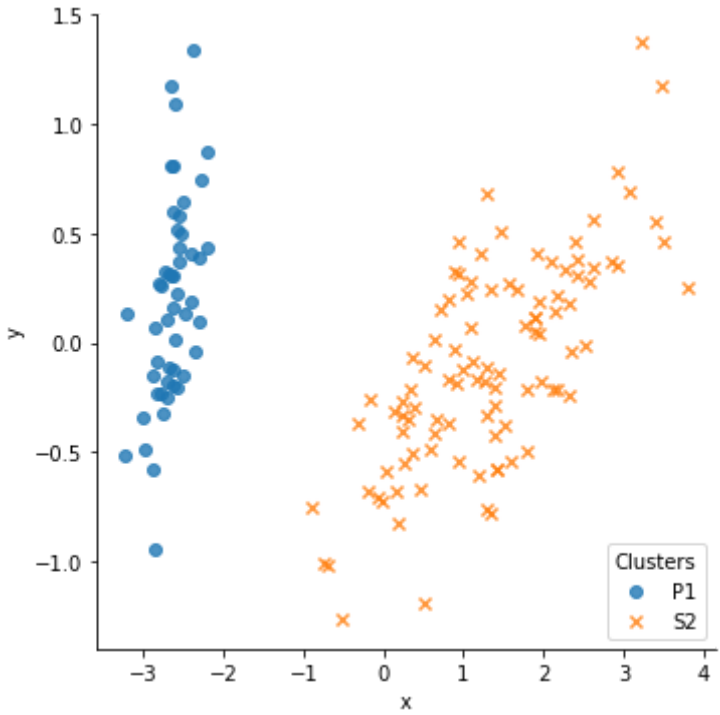

*Fig. 2: 3EC with input cluster A and $\varphi_b$ - μ yields **Agglo - 2***

3EC further partitions S2 and the **Γ** function outputs the $\varphi_b$ - μ pair as Agglomerative and 2 respectively. No further partitioning is possible because it breaches the threshold of the length of individual partitions (β).

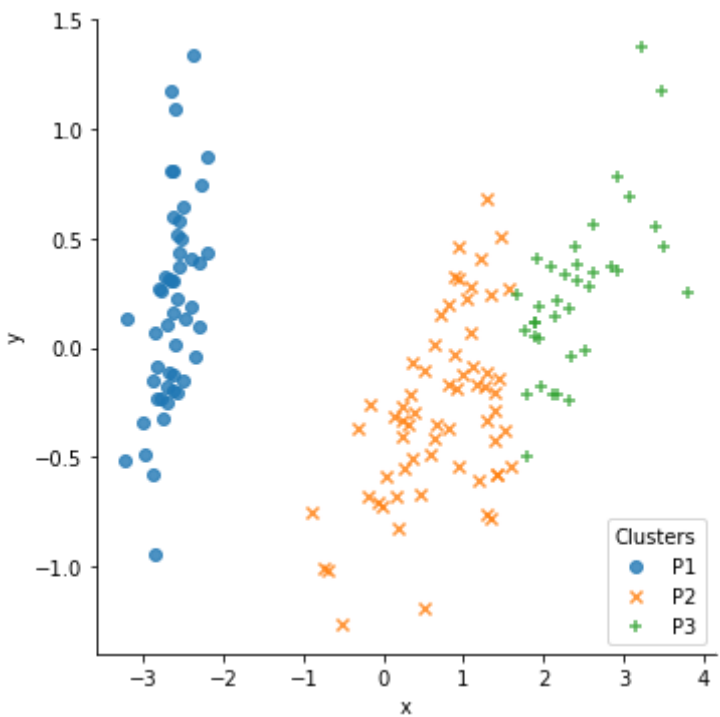

*Fig. 3: 3EC with input cluster S2 and $\varphi_b$ - μ yields **Agglo - 2***

## 10.3 Observations:

(1) The data is linearly separable between two clusters and that is evident in Fig. 2. This partitioning is easy.

(2) The partitioning gets difficult when we split S2 because it is not linearly separable and the points are in very close proximity.

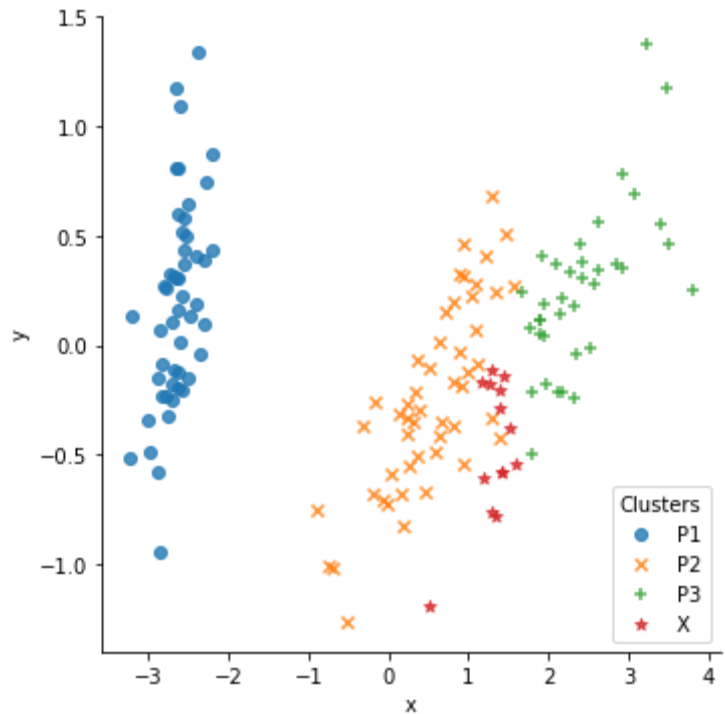

*Fig. 4: The cluster X does not conform with the labelled data*

It is not the intent to be influenced by the labelled data, but if we were to compare the outputs then the data points of the 4th cluster of Fig. 4 are the data points that are not classified as per the labels. It is evident that the data points are in such close proximity that only a domain expert can provide the view point on the right classification. Conversely, if the data has not been classified then the different cluster outputs based on variations of the input criteria, helps the business domain expert to come up with the right rationale for classification.

(3) This is a very simple case study with only 150 x 4 data points.
(4) The stopping criteria is an effective tool to control the number of splits and the quality of the individual splits
(5) Since this is a data discovery process, the final discretion is with the data scientist and the domain expert:
    (a) Which algorithms to choose?
    (b) Which clustering validation indices to choose?
    (c) What stopping criteria to choose?

# 11. Conclusion

The 3EC algorithm is in essence a data exploration framework that allows the identification of new clusters hitherto unknown in a new or a refreshed data set. The Tau grid is a score based evaluation of the quality of clusters that results from the partitioning unique to individual algorithms of an ensemble of algorithms. It requires both a data scientist and a domain expert to identify the new cluster patterns. The creation of the Tau grid is compute and time intensive but it is a one time effort for a given set of data and input configuration.